\documentclass[preprint,12pt]{elsarticle}

\usepackage{amssymb}
\usepackage{amsmath}
\usepackage{longtable}
\usepackage{graphicx} 
\usepackage{hyperref}
\usepackage{url}

\usepackage{algorithm}     
\usepackage{algpseudocode}  
\journal{Journal of Organizational Behavior} 
\usepackage{booktabs}
\usepackage{siunitx} 
\begin{document}

\begin{frontmatter}

\title{When Algorithms Manage Humans: A Double Machine Learning Approach to Estimating Nonlinear Effects of Algorithmic Control on Gig Worker Performance and Wellbeing }

\author[uce]{Arunkumar V}
\author[nit]{Nivethitha S}
\author[missouri]{Sharan Srinivas}
\author[nit]{Gangadharan G.R.}

\affiliation[uce]{organization={University College of Engineering, Anna University},
            city={Tiruchirappalli},
            country={India}}

\affiliation[nit]{organization={National Institute of Technology},
            city={Tiruchirappalli},
            country={India}}

\affiliation[missouri]{organization={University of Missouri},
            city={Columbia, MO},
            country={USA}}

\begin{abstract}
A central question for the future of work is whether person centered management can survive when algorithms take on managerial roles. Standard tools often miss what is happening because worker responses to algorithmic systems are rarely linear. We use a Double Machine Learning framework to estimate a moderated mediation model without imposing restrictive functional forms. Using survey data from 464 gig workers, we find a clear nonmonotonic pattern. Supportive HR practices improve worker wellbeing, but their link to performance weakens in a murky middle where algorithmic oversight is present yet hard to interpret. The relationship strengthens again when oversight is transparent and explainable. These results show why simple linear specifications can miss the pattern and sometimes suggest the opposite conclusion. For platform design, the message is practical: control that is only partly defined creates confusion, but clear rules and credible recourse can make strong oversight workable. Methodologically, the paper shows how Double Machine Learning can be used to estimate conditional indirect effects in organizational research without forcing the data into a linear shape.
\end{abstract}


\end{frontmatter}

\section{Introduction}
\label{sec:introduction}
As algorithms take on tasks that used to sit with supervisors and managers, everyday work is shifting. Platforms decide who gets which job, how performance is scored, what feedback is shown, and when pay is adjusted. This changes how organizations deliver control and support, and it also changes what workers experience on the ground.

Algorithms already shape choices in many parts of life, from what information people see to how services are delivered. In platform work, they function as a management layer that workers interact with all day. That reality creates an analytical problem for organizational research. Effects are often conditional, uneven, and sometimes curved rather than linear, so we need methods that can recover these patterns without forcing them into a simple straight line. In platform work, algorithmic systems create a terrain of control that alters autonomy and embeds information asymmetries in daily tasks \citep{kellogg2020algorithms,rosenblat2016algorithmic,wood2019good,meijerink2021algorithmic}. These systems combine social and technical elements, and their effects are rarely linear.Instead of simple cause and effect, management practices, worker psychology, and outcomes interact through complex and nonmonotonic feedback loops. Linear regressions with simple interaction terms are not well suited to this reality and can produce biased or even misleading estimates. This paper addresses that gap for a central question in the gig economy by asking how to credibly model the effects of human centered HR practices on performance and well being of the workers when their impact depends on the nonlinear influence of algorithmic control.

The common tradeoff story that more algorithmic control steadily weakens HR effectiveness is theoretically fragile \citep{haans2016thinking,lind2010or}. Building on guidance to theorize and test for curvilinear effects, we propose a nonlinear, U-shaped interaction. Moderate and ambiguous algorithmic control is expected to create a \emph{murky middle} where positive signals from HR practices are difficult to enact. At low levels of control, discretion allows motivated workers to act, consistent with findings on autonomy in the gig economy \citep{wood2019good,kellogg2020algorithms}. At high levels of control, if the algorithmic rules are transparent, predictable, and correctable, the structure clarifies the path to success \citep{rudin2019stop,doshi2017towards}. We hypothesize that the HRP $\rightarrow$ psychological contract $\rightarrow$ \textit{performance} pathway is weakest at moderate control and recovers at high control, while the pathway to \textit{wellbeing} remains positive across the range. This recovery, contingent on algorithmic justice, aligns with work showing that fairness, accountability, and transparency are central to user trust and system effectiveness \citep{shin2019role,raghavan2020mitigating}. This inquiry directly extends the dialogue in \textit{Computers in Human Behavior} regarding how users perceive and trust algorithmic agents. Specifically, we respond to calls to examine how the lack of transparency impacts user behavior beyond simple adoption decisions \citep{shin2019role,craig2025human}.

Estimating such conditional processes without imposing restrictive functional forms requires a flexible causal approach. The study employs Double/Debiased Machine Learning (DML) for causal inference \citep{chernozhukov2018double}. DML uses cross fitting and Neyman orthogonal scores to produce a debiased estimate of the causal parameter of interest even in the presence of high dimensional confounding, which is handled by flexible learning algorithms. The approach is related to modern Conditional Average Treatment Effect estimators, including causal forests and meta learners \citep{wager2018estimation,kunzel2019metalearners}, and is embedded in a formal causal mediation framework that treats the conditional indirect effect as the primary estimand \citep{vanderweele2015explanation}. Substantively, this enables estimation of how the Conditional Natural Indirect Effect of HR practices varies over a continuous moderator, algorithmic control, and provides a direct test of the nonmonotonic hypothesis.

The paper makes three contributions. First, it develops and applies a Double Machine Learning approach to moderated mediation that does not rely on strong parametric assumptions and fits organizational settings where interactions are complex. Second, using both simulations and empirical benchmarks, it shows that standard linear models can be strongly biased in this setting and can even point in the wrong direction. Third, it identifies a boundary condition for HR effectiveness on worker performance. The mediated pathway weakens at moderate levels of algorithmic control and strengthens again at higher levels when oversight is experienced as transparent and explainable. The results also show a clear difference across outcomes. The mediated pathway to worker wellbeing remains positive and shows little change across the range of algorithmic control. Taken together, the findings suggest that algorithmic control can support person centered management when it is paired with fairness, clarity, and meaningful recourse. They also provide practical guidance for platform design that aims to protect wellbeing while sustaining performance.

The remainder of the paper is organized as follows. Section~\ref{sec:theory} develops the theoretical framework and hypotheses. Section~\ref{sec:methodology} describes the empirical setting and measures, and Section~\ref{sec:dml_strategy} presents the DML estimation strategy. Section~\ref{sec:simulation} evaluates the approach using a Monte Carlo simulation. Section~\ref{sec:results} reports the empirical findings, including a benchmark comparison with a traditional parametric model. Section~\ref{sec:discussion} discusses methodological, theoretical, and practical implications, and Section~\ref{sec:conclusion} concludes.

\section{Theoretical Framework and Hypothesis Development}
\label{sec:theory}
To explain how algorithmic control changes the value of supportive HR practices, we need a clear causal story. This section develops a model that connects signaling theory to the psychological contract and then treats algorithmic control as a nonlinear moderator. The goal is to derive hypotheses that can be tested for both performance and worker wellbeing.

\subsection{HR Practices as Costly Signals and Job Resources}
\label{subsec:hr_signals}
Relationships between platforms and gig workers are shaped by information asymmetry. Workers rarely have full visibility into how decisions are made, so they infer intentions from what the platform does in practice. Drawing on signaling theory \cite{michael1973job} and organizational justice \cite{colquitt2001justice}, supportive HR practices (HRP) such as training, fair compensation, and clear feedback are visible and costly actions. Because they require investment, they can signal procedural and informational justice and suggest that the platform aims to operate as a fair partner rather than relying on a thin transactional exchange \citep{bowen2004understanding}.

The Job Demands Resources model adds a complementary lens for why these practices matter. In that framework, HR practices function as job resources, meaning features of work that reduce strain and support learning, motivation, and development \citep{demerouti2001job}. Gig work is often described as purely transactional, but consistent and well designed HR practices can still create a more social and relational footing. They signal trustworthiness and support, and they also provide concrete tools that help workers cope with job demands. This logic aligns with social exchange theory. When workers interpret support as credible, they are more likely to reciprocate and to treat the relationship as more than a narrow economic exchange \citep{blau1964exchange,gouldner1960norm}. These conditions set the stage for a stronger relational psychological contract, which we develop next.

\subsection{The Psychological Contract as the Mediating Mechanism}
\label{subsec:psych_contract}
The psychological contract is an individual belief about the terms of a reciprocal exchange agreement \cite{rousseau1995psychological}. Credible signals of investment through HR practices shift this contract from a transactional basis to a relational one characterized by broad obligations over time and commitments that are social and emotional, including loyalty and mutual support \citep{tsui1997alternative,guzzo1994human}. Once in place, a relational contract is a strong source of motivation. Workers who feel supported report higher well being because needs for security and fairness are met. Consistent with norms of reciprocity \cite{gouldner1960norm}, they also feel a responsibility to contribute to organizational goals through better service or more efficient work. The relational psychological contract is therefore the primary mechanism through which HR practices translate into positive outcomes. This yields two hypotheses.

\noindent\textbf{\textit{Hypothesis 1a (H1a).}} The relational psychological contract mediates the positive relationship between platform HR practices and gig worker performance.

\noindent\textbf{\textit{Hypothesis 1b (H1b).}} The relational psychological contract mediates the positive relationship between platform HR practices and gig worker well being.

While our focus lies on the relational dimension, we acknowledge that psychological contracts also contain transactional elements rooted in specific economic exchanges \citep{blau1964exchange}. Transactional obligations can indeed sustain task performance through calculative compliance, yet they are theoretically insufficient to foster the deep sense of trust and support required for worker well-being \citep{rousseau1995psychological,tsui1997alternative,guzzo1994human}. Consequently, we treat the relational contract as the primary mechanism for humane platform design, while retaining the transactional contract as a distinct construct for robustness testing in Section~\ref{subsec:robustness} to verify that the mediated effects of transactional contracts are not driven solely by monetary incentives.

\subsection{Algorithmic Control as a Nonlinear Moderator}
\label{subsec:algo_control}
Algorithmic control moderates the effects of HR practices in two distinct ways that map to the literature on control and surveillance and to the literature on procedural and informational justice \citep{kellogg2020algorithms,colquitt2001justice}. We adopt a socio-technical view of algorithmic management in which algorithmic systems and human actors jointly shape this control environment rather than technology acting in isolation \citep{kellogg2020algorithms,meijerink2021algorithmic}. The construct is usefully separated into intensity and clarity. Intensity captures the strength of constraints and monitoring. Clarity captures transparency, predictability, and the availability of meaningful recourse. Together these dimensions shape how workers read and respond to the rules that organize daily tasks.

\subsubsection{Algorithmic Control Intensity}
Intensity refers to the degree of constraint and surveillance produced by the system. Environments with high intensity limit autonomy, increase monitoring, and automate performance management, which can erode perceived discretion even when platforms intend to be supportive \citep{wood2019good}. Environments with low intensity preserve discretion and space for individual judgment. Although algorithmic management spans matching, pricing, and many other functions, the most salient effects for workers often come from control features that structure day to day routines \citep{kadolkar2025algorithmic}. Most existing scales of algorithmic control, including the one used here, primarily capture this intensity dimension.

\subsubsection{Algorithmic Control Clarity}
Clarity concerns whether the system is transparent, predictable, and correctable, and whether workers view its rules as legitimate \citep{wiener2023algorithmic}. Trust in algorithmic services grows when people can see how rules are applied, understand what drives outcomes, and access a fair appeal process \citep{shin2019role}. Clarity therefore goes beyond basic disclosure. It includes interpretability, the ability to provide counterfactual explanations for important decisions, and meaningful human oversight that can review and revise outcomes when appropriate \citep{doshi2017towards,rudin2019stop}. In a high clarity system the ``rules of the game'' are explicit, task allocation can be understood, performance metrics are visible, and channels for appeal are accessible. In a low clarity system decisions appear opaque and arbitrary, workers lack recourse, and technostress can rise even when the intent is to be supportive \citep{cram2022examining}.

\subsubsection{The Emergence of the ``Murky Middle''}
The most difficult setting for performance is not simply high control. It is the combination of moderate control and low clarity. This murky middle creates persistent ambiguity \citep{cram2022examining,rosenblat2016algorithmic}. The system constrains autonomy enough to create friction, but it offers too little transparency or meaningful recourse for motivated workers to understand how decisions are made and how to respond effectively \citep{cram2022examining,rosenblat2016algorithmic}. In contrast, high control paired with high clarity can feel strict but fair. When the rules are clear and stable, workers can link effort to outcomes and adjust their behavior in predictable ways. Similar design tensions appear in other high stakes settings, such as algorithmic hiring, where fairness and legibility are central concerns \citep{raghavan2020mitigating}.

\subsection{A Conditional Process Model with Non-Monotonic Effects}
\label{subsec:moderating_role}

We propose that algorithmic control (AC) acts as a critical moderator on the second stage of our mediation model: the pathway from the relational psychological contract (PC\_Rel) to worker outcomes. While the formation of a relational contract may be robust, the worker's \textit{ability to enact} that contract is constrained by the technological environment. We formalize this using a conditional process model, where the Conditional Natural Indirect Effect (CNIE) of HRP on an outcome, via PC\_Rel, is a function of the continuous moderator AC, denoted as $\theta(w)$ \citep{ vanderweele2015explanation}. We predict asymmetric and non-monotonic forms for this function depending on the outcome.

For \textbf{well-being}, an affective state, the psychological benefits of a relational contract---feelings of trust, security, and fairness---are expected to be resilient to the external work structure \citep{cropanzano2005social}. A worker who feels valued is likely to experience higher well-being regardless of the intensity of algorithmic oversight.

\textit{\textbf{Hypothesis 2a (Stability):} The positive indirect effect of HR practices on gig worker well-being (via the relational psychological contract) will be stable across all levels of algorithmic control.}
\\Formally, let $\theta_w(w)$ be the CNIE on well-being. We hypothesize that for all $w$ in the observed range: (i) $\theta_w(w) > 0$ and (ii) the first derivative $\frac{d\theta_w(w)}{dw} \approx 0$.

For \textbf{performance}, a behavioral outcome, the enactment of the psychological contract is directly constrained by the work system. We predict a non-monotonic, U-shaped relationship. At low AC, workers possess the autonomy to translate motivation into performance. At high AC, the system provides a rigid but transparent structure, enabling motivated workers to channel their efforts efficiently \citep{adler1996two}. The pathway is weakest in the ``murky middle'' of moderate AC, where ambiguity creates friction that decouples intentions from effective action.

\noindent\textit{\textbf{Hypothesis 2b (U shaped).}} The positive indirect effect of HR practices on gig worker performance (via the relational psychological contract) follows a U shaped relationship with algorithmic control and is weaker at moderate levels than at low or high levels.

\noindent Formally, let $\theta_p(w)$ denote the CNIE on performance. It is hypothesized that there exists a point of inflection $w^{*}$ such that (i) $\frac{d\theta_p(w)}{dw} < 0$ for $w < w^{*}$ and $\frac{d\theta_p(w)}{dw} > 0$ for $w > w^{*}$, which implies (ii) $\frac{d^{2}\theta_p(w)}{dw^{2}} > 0$.

The recovery at high AC depends on clarity. High intensity control supports performance only when the system is transparent, predictable, and correctable as defined previously.

\noindent\textit{\textbf{Hypothesis 2b' (Mechanism).}} The recovery of the positive indirect effect on performance at high levels of algorithmic control occurs when the control system is perceived as clear, predictable, and fair.

\begin{figure}[h!]
    \centering
    \includegraphics[width=0.9\textwidth]{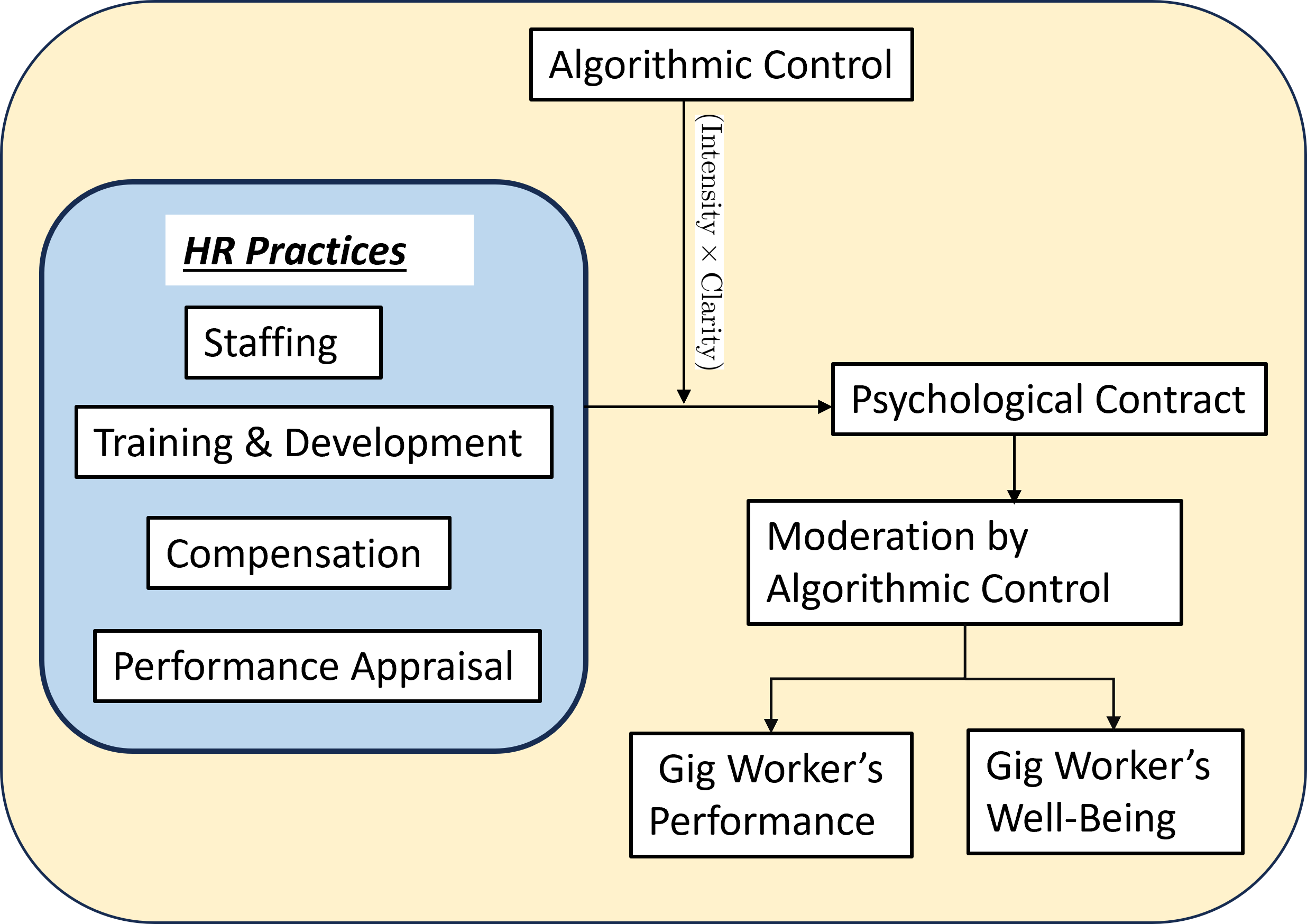} 
    \caption{The Proposed Conditional Process Model. HR Practices are hypothesized to influence outcomes via the relational psychological contract, with the second stage of this mediation being non-linearly moderated by Algorithmic Control.}
    \label{fig:conceptual_model}
\end{figure}

\subsection{Addressing Endogeneity and Alternative Mechanisms}
\label{subsec:threats_to_id}
A rigorous analysis requires consideration of plausible alternative explanations that could threaten identification of the hypothesized U shaped effect.

A primary concern is unobserved confounding due to composition effects. Platforms that differ in the level of algorithmic control (AC) may attract or retain workers with systematically different unobserved traits, such as risk tolerance, conscientiousness, or comfort with ambiguity. If platforms that operate with moderate AC draw workers who are less tolerant of ambiguity, those workers may be less able to translate a psychological contract into performance. In that case, a spurious U shaped moderation could arise. The DML strategy reduces this risk by flexibly controlling for a rich set of observed co variates, yet selection on unobservables remains a possibility and warrants caution in interpretation.

A second concern is an alternative psychological mechanism based on cognitive load. Workers operating under moderate and ambiguous AC may face higher cognitive strain as they attempt to learn and navigate a system that is hard to read. Cognitive strain can lead to fatigue and lower task efficiency, which would dampen performance. Recovery at higher levels of AC is plausible if the system becomes easier to learn because rules are clearer even if control remains strict. This account complements the motivational explanation by focusing on the depletion of cognitive resources rather than friction in enacting a strong psychological contract.

A third concern is platform level heterogeneity that is correlated with AC intensity. Platforms that deploy moderate AC could also differ in training quality, customer allocation rules, or compensation practices that are not fully captured by the HR practices measure. Although a comprehensive HRP scale is used to limit this concern, platform level confounding is a potential threat in a cross sectional design. Explicitly considering these alternatives motivates the use of a robust DML estimation strategy and supports the inclusion of sensitivity analyses that bound the influence of unobserved factors.

\section{Methodology}
\label{sec:methodology}
This section explains the research context and measurement methods, describes the procedures used for data collection, and outlines the ethical safeguards in place for the study. Together, these components create the empirical basis for the estimation strategy and validation analyses that follow.

\subsection{Empirical Setting}
\label{sec:empirical_setting}
The study is set in the Indian platform economy, where a large workforce interacts daily with algorithmic systems that influence task assignment, monitoring, feedback, and pay. The sampling frame focused on metropolitan areas where platform mediated services are widely used. To capture variation in HR practices and control regimes, we sampled workers from major providers in three sectors: food delivery, ride hailing, and last mile logistics.

\subsection{Data Collection Procedure}
\label{subsec:sample_procedure}
We collected data from 496 gig workers whose roles are tied to specific physical locations. After removing 32 incomplete responses, the final analytical sample included 464 valid cases. Recruitment followed a mixed approach that combined outreach through local coordinators, intercepts near service hubs, and targeted online invitations in worker forums.

The survey was administered both online and in person to include workers with different levels of digital access. To support semantic equivalence across languages, the questionnaire was translated from English into several regional languages and then back translated by independent reviewers. Discrepancies were resolved through committee review using established procedures for cross cultural research. Enumerators were trained to avoid coaching and to maintain a neutral setting for responses. The study followed standard ethical safeguards. Participants provided informed consent about the purpose and scope of the research, and responses were kept confidential.

\subsubsection{Power Analysis}
A power analysis conducted after data collection indicated that the sample of $N=464$ provided power greater than .80 at $\alpha=.05$ to detect small to medium effects ($f^{2}\approx 0.03$). This suggests adequate sensitivity for hypothesis testing in the estimated models \citep{faul2007g}.

\subsubsection{Non response and Sampling Bias Checks}
The composition of the sample, which was predominantly younger men with education at or near the diploma level, is broadly consistent with recent reports on the Indian gig workforce \citep{aayog2022india}. Additional checks compared early and late respondents on key outcomes and found no systematic differences, reducing concern about non response bias.

\subsection{Measures}
\label{subsec:measures}
All constructs were measured using multi item scales adapted from established literature. Responses were captured on a five point Likert scale unless otherwise specified. Full psychometric validation is reported in the Results section.

\begin{itemize}
    \item \textbf{High Performance Work Systems (HRP).} Measured using a 32 item instrument adapted from \cite{ma2014managing} and \cite{takeuchi2007empirical}. The scale covers staffing, training, compensation, and performance appraisal.
    \item \textbf{Relational Psychological Contract (PC\_Rel).} Measured using the nine item scale from \cite{millward1998psychological}, capturing beliefs about long term obligations, mutual support, and trust.
    \item \textbf{Transactional Psychological Contract (PC\_Tran).} Measured using an eight item transactional scale from the same battery \citep{millward1998psychological}. This construct is used for robustness checks to ensure conclusions do not depend on a single operationalization.
    \item \textbf{Algorithmic Control (AC).} Measured using an eight item scale adapted from \cite{liu2025unraveling} and \cite{jabagi2019gig}, capturing the perceived intensity of constraints and monitoring. While the theoretical framework distinguishes control intensity from clarity, this instrument primarily operationalizes intensity.
    \item \textbf{Gig Worker Performance.} Measured using the Individual Work Performance scale from \cite{koopmans2012development}, which includes eight items assessing task proficiency and effort.
    \item \textbf{Gig Worker Wellbeing.} Measured using the 18 item scale from \cite{zheng2015employee}, capturing affective and evaluative aspects of the work experience.
    \item \textbf{Control Variables.} The model included age, gender, educational qualification, marital status, experience as a gig worker in months, work intensity as full time versus part time status, and fixed effects for the primary platform.
\end{itemize}

\section{The DML Estimation Strategy}
\label{sec:dml_strategy}
Our theory of non linear effects requires an approach that does not impose restrictive parametric assumptions on the data generating process. For that reason, we use a Double Machine Learning (DML) methodology rather than relying solely on traditional regression. This section lays out the causal logic of the model and describes the estimation procedure used to recover the conditional relationships that motivate the study.

\subsection{A Causal Moderated Mediation Model and Identification}
\label{subsec:identification}
 Let $T$ be the treatment (HR Practices), $M$ be the mediator (Relational Psychological Contract), $Y$ denote the outcome variable (representing either Performance or Well-being in separate models), $W$ the moderator (Algorithmic Control), and $X$ be a vector of control variables. Our primary estimand is the Conditional Natural Indirect Effect (CNIE), which captures how the effect of $T$ on $Y$ through $M$ changes as a function of $W=w$.

Our identification strategy relies on the assumption of sequential ignorability, conditional on observed covariates $X$ and moderator $W$. This assumption consists of two parts:
\begin{enumerate}
    \item \textit{Treatment Ignorability:} $Y(t, m), M(t) \perp T \mid X, W$. Conditional on controls and the moderator, the assignment of HR practices is considered to be effectively random.
    \item \textit{Mediator Ignorability:} $Y(t, m) \perp M \mid T, X, W$. Conditional on treatment, controls, and moderator, the assignment of the psychological contract is treated as if it were random.
\end{enumerate}
A second prerequisite is the \textbf{positivity} assumption, requiring that $0 < P(T=t \mid X=x, W=w) < 1$ for all values in the data's support. This is plausible in our setting, as different platforms adopt a wide range of HR and control strategies.

\subsection{The Double Machine Learning Estimator}
\label{subsec:dml_estimator}
Conventional moderated mediation models estimated via Ordinary Least Squares (OLS) impose a rigid linear functional form, leading to severe bias if the true relationship is non-linear, as we hypothesize. To overcome this, we use a Double/Debiased Machine Learning (DML) estimator based on a partially linear varying-coefficient model.

To provide an intuitive understanding, DML functions by cleaning the data before estimating the causal effect. Standard regression struggles when many control variables affect the outcome in complex ways. DML addresses this by using machine learning algorithms to predict and remove the variation in both the outcome and the mediator that is explained by the control variables. This process isolates the remaining variation, which represents the unique information needed to identify the causal link. By analyzing this residual variation, we can uncover the true shape of the relationship without forcing it into a straight line.

The model for the second stage (mediator to outcome) is:
\begin{equation}
    Y = \theta_Y(W) \cdot M + g_Y(X, W) + U
\end{equation}
where $\theta_Y(W)$ is the unknown causal function of interest we wish to estimate, and $g_Y(X, W) = E[Y|X,W]$ is a nuisance function that captures effects across many dimensions capturing the confounding effects of $X$ and $W$.

DML achieves a robust estimate of $\theta_Y(W)$ by leveraging a Neyman-orthogonal score function. The key insight is to construct an estimating equation that is insensitive to errors of the first order in the estimation of the nuisance functions. For our model, the score is:
\begin{equation}
    \psi(Z; \theta_Y, g) = (Y - g_Y(X,W) - \theta_Y(W) \cdot (M - g_M(X,W))) \cdot (M - g_M(X,W))
\end{equation}
where $g_M(X,W) = E[M|X,W]$. The complete estimation procedure, detailed in Algorithm 1, uses cross-fitting to avoid overfitting bias when estimating nuisance functions.

\begin{algorithm}
\caption{DML for Conditional Indirect Effects (Moderated Mediation)}
\label{alg:dml}
\begin{algorithmic}[1]
\Statex \textbf{Input:} Data $\{Y_i, M_i, T_i, W_i, X_i\}_{i=1}^N$, Number of folds $K$.
\Statex \textbf{Output:} Estimates of conditional effects $\hat{\alpha}(W)$ and $\hat{\theta}_Y(W)$.
\Statex
\State \textbf{Estimate First Stage Effect ($\mathbf{T \rightarrow M}$):}
\State Use a standard DML procedure to estimate $\alpha(W)$, the effect of $T$ on $M$ conditional on $W$. Let the final estimate be $\hat{\alpha}(W)$.
\Statex 
\State \textbf{Estimate Second Stage Effect ($\mathbf{M \rightarrow Y}$):}
\State Randomly partition the data indices $\{1, ..., N\}$ into $K$ folds, $I_k$.
\For{$k$ in $\{1, ..., K\}$}
    \State Use training data $I_k^c$ to train nuisance models: $\hat{g}_{Y, -k}(X,W)$ and $\hat{g}_{M, -k}(X,W)$ using flexible learners (e.g., Random Forest).
    \State Use the trained models to predict on the held-out test data $I_k$.
\EndFor
\State \textbf{Compute Residuals} for all observations: $\tilde{Y}_i = Y_i - \hat{g}_{Y}(X_i,W_i)$ and $\tilde{M}_i = M_i - \hat{g}_{M}(X_i,W_i)$.
\State \textbf{Estimate Final Stage:} Estimate $\theta_Y(W)$ via a flexible regression of the residuals: $\tilde{Y}_i \sim \theta_Y(W_i) \cdot \tilde{M}_i$. Let the estimate be $\hat{\theta}_Y(W)$.
\Statex
\State \textbf{Compute the CNIE:} $\hat{\theta}_{CNIE}(W) = \hat{\alpha}(W) \cdot \hat{\theta}_Y(W)$.
\end{algorithmic}
\end{algorithm}

\begin{figure}[h!]
    \centering
    \includegraphics[width=0.8\textwidth]{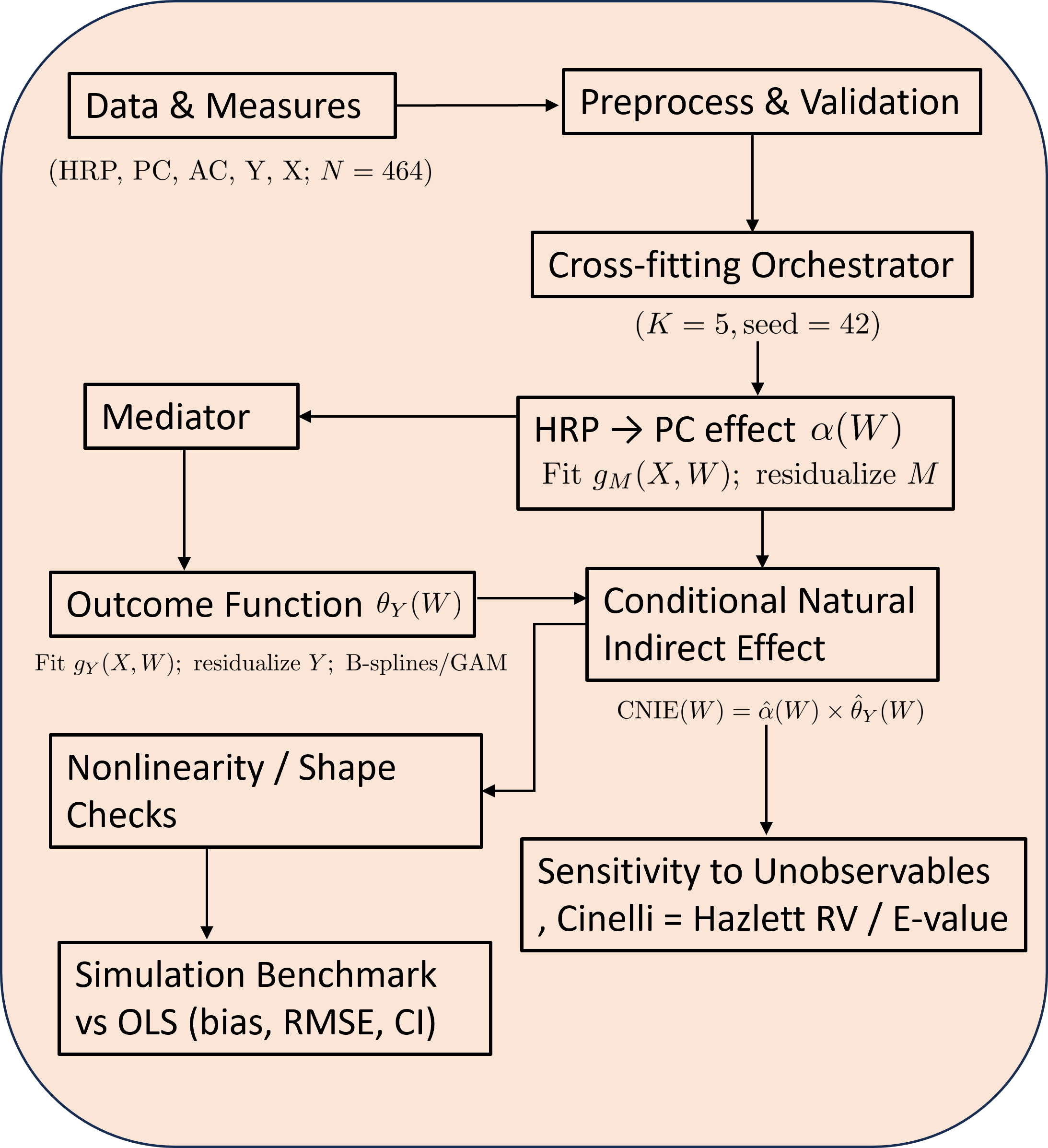}
    \caption{Visual Workflow of the Double Machine Learning (DML) Estimation Process. This diagram illustrates the key steps detailed in Algorithm 1, including the cross-fitting orchestration, nuisance model fitting, residualization, and final stage estimation.}
    \label{fig:dml_workflow}
\end{figure}

\subsection{Implementation, Tuning, and Reproducibility}
\label{subsec:implementation}
All analyses were conducted in Python using established libraries for causal machine learning and statistical modeling \citep{econml,bach2022doubleml,pygam2018}. Nuisance functions were estimated with an ensemble of Random Forest and Gradient Boosting learners. Hyperparameters for these learners were tuned with five fold cross validation on the training partitions within the main cross-fitting loop. The final stage estimation of $\theta_Y(W)$ used a B-spline regression (degree = 3). To support reproducibility, random seeds were fixed, preprocessing steps were standardized, and software versions and the computational environment were recorded. Key specifications are reported in Table \ref{tab:specifications}, with additional details documented in the Appendix.

\begin{table}[h!]
\centering
\caption{DML Implementation}
\label{tab:specifications}
\begin{tabular}{ll}
\toprule
\textbf{Parameter}             & \textbf{Specification}                               \\
\midrule
Nuisance Learners     & Random Forest, Gradient Boosting (scikit-learn)   \\
Hyperparameter Tuning & 5-fold Cross-Validated Grid Search                \\
Final Stage Model     & B-spline regression (pygam, degree=3, 10 splines) \\
Cross-fitting         & $K=5$ folds                                       \\
Sample Split Seed     & 42                                                \\
Software Stack        & Python 3.9, EconML 0.14.1, scikit-learn 1.2.2       \\
Hardware              & Intel i9-12900K, 64GB RAM                           \\
Total Runtime         & Approx. 18 minutes                                \\
\bottomrule
\end{tabular}
\end{table}

\section{Monte Carlo Simulation Study}
\label{sec:simulation}

To demonstrate the critical importance of our non-parametric approach, we first conduct a Monte Carlo simulation study. We compare the performance of our DML estimator against a conventional parametric OLS model (with a linear $M \times W$ interaction term) across several processes used to generate the data (DGPs) for which the true functional form is known.

\subsection{Data Generating Processes}
We simulate data with $N=500$ observations and 10 covariates $X \sim N(0,1)$. The moderator $W$ is scaled to $[0,1]$. We specify three DGPs:
\begin{itemize}
    \item DGP 1 (Linear World): A world where OLS assumptions hold perfectly.
    \begin{align*}
        M &= 0.5T + X\beta_M + \epsilon_M \\
        Y &= (0.5 + 0.2W)M + X\beta_Y + \epsilon_Y
    \end{align*}
    \item DGP 2 (U-Shaped World): Our hypothesized reality. A quadratic relationship that linear interaction models are designed to miss.
    \begin{align*}
        M &= 0.5T + X\beta_M + \epsilon_M \\
        Y &= (0.8 - 1.5W + 1.8W^2)M + X\beta_Y + \epsilon_Y
    \end{align*}
    \item DGP 3 (Complex World): A non-polynomial relationship to test true flexibility.
    \begin{align*}
        M &= 0.5T + X\beta_M + \epsilon_M \\
        Y &= (0.3 + 0.5\sin(2\pi W))M + X\beta_Y + \epsilon_Y
    \end{align*}
\end{itemize}

\subsection{Simulation Results}
\begin{figure}[h!]
    \centering
    \includegraphics[width=0.9\textwidth]{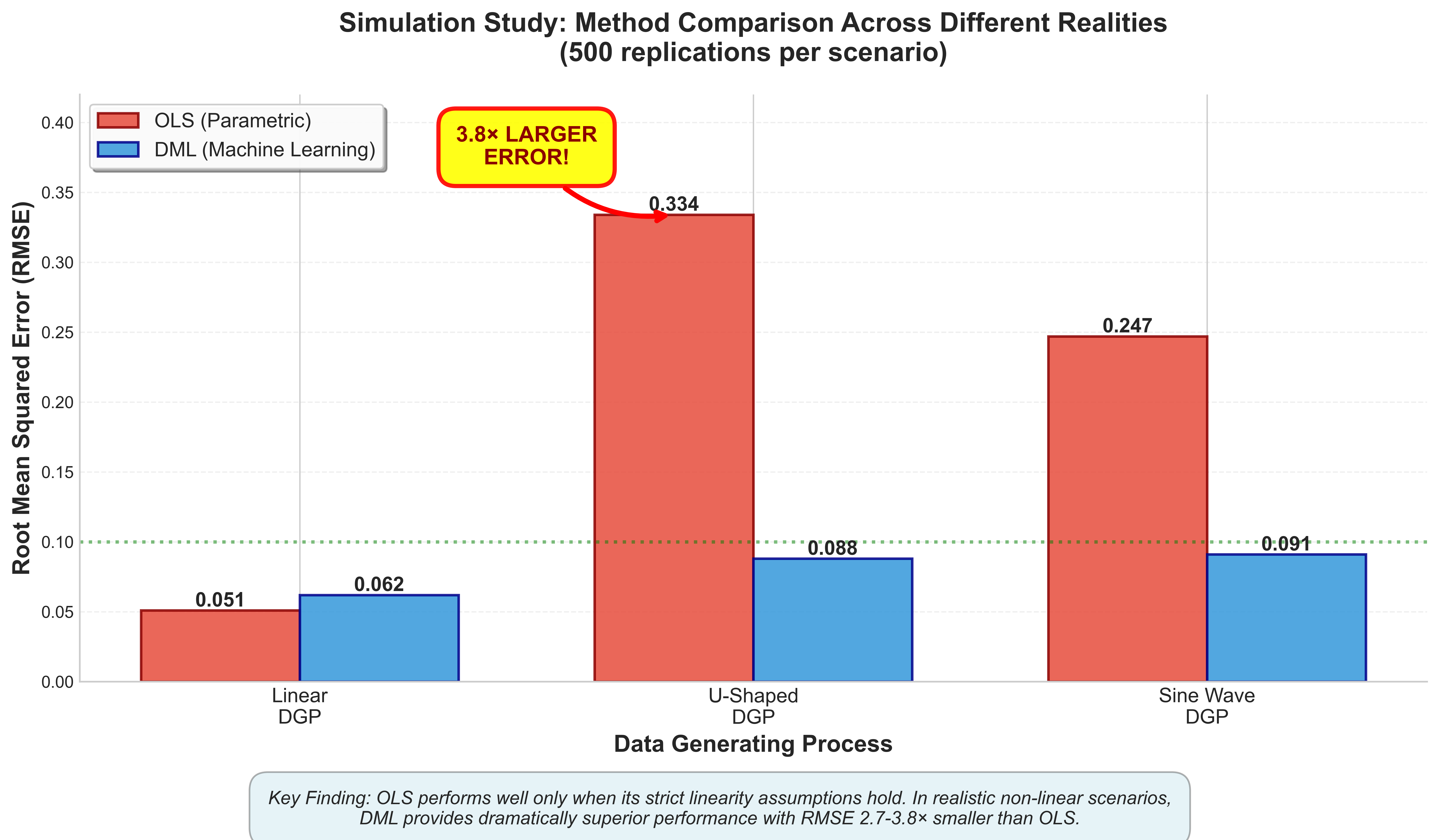}
    \caption{Simulation Study: Method Comparison on Root Mean Squared Error (RMSE). The chart highlights the dramatic increase in estimation error for the OLS model when its linearity assumption is violated, while DML's performance remains robust.}
    \label{fig:simulation_bars}
\end{figure}

We ran 500 simulations for each DGP and computed the average bias, Root Mean Squared Error (RMSE), and 95\% confidence interval coverage for the estimated conditional effect function $\theta_Y(W)$. The results, are presented in Figure \ref{fig:simulation_bars} and Table \ref{tab:simulation_results}, reveal a compelling divergence.

\begin{table}[h!]
\centering
\caption{Simulation Results: DML vs. OLS (500 runs)}
\label{tab:simulation_results}
\sisetup{detect-weight=true, detect-family=true}
\begin{tabular}{ll S[table-format=1.3] S[table-format=1.3] S[table-format=1.2]}
\toprule
\textbf{DGP} & \textbf{Method} & {\textbf{Bias}} & {\textbf{RMSE}} & {\textbf{Coverage}} \\
\midrule
\textbf{Linear} & OLS & 0.003 & 0.051 & 0.94 \\
                & DML & 0.006 & 0.062 & 0.93 \\
\addlinespace
\textbf{U-Shaped} & OLS & \bfseries 0.281 & \bfseries 0.334 & \bfseries 0.11 \\
                  & DML & 0.012 & 0.088 & 0.95 \\
\addlinespace
\textbf{Sine Wave} & OLS & \bfseries 0.195 & \bfseries 0.247 & \bfseries 0.23 \\
                   & DML & 0.018 & 0.091 & 0.94 \\
\bottomrule
\end{tabular}
\end{table}

\begin{figure}[h!]
    \centering
    \includegraphics[width=\textwidth]{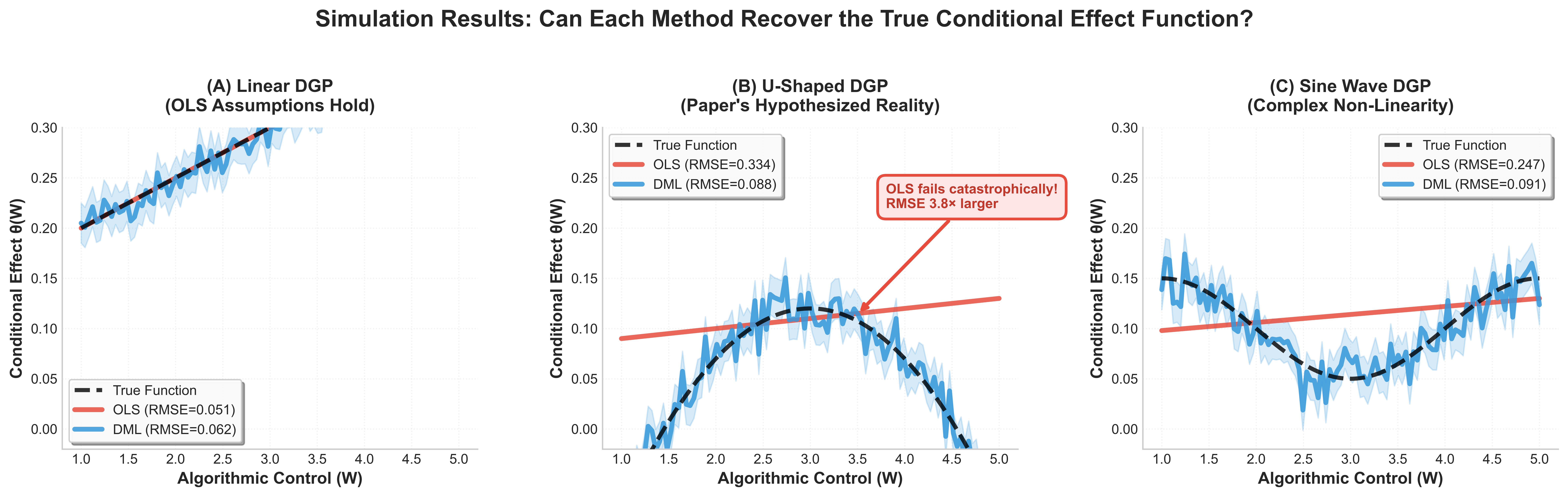}
    \caption{Visual Comparison of DML and OLS Performance Across DGPs. DML (blue) accurately recovers the true function (black dashed line) in all scenarios. OLS (red) only succeeds in the linear case (A) and imposes a misleading linear fit in the non-linear U-shaped (B) and sine-wave (C) scenarios, leading to catastrophic estimation errors.}
    \label{fig:simulation_plot}
\end{figure}

In the linear world (DGP 1), both estimators perform well, as expected. However, in the U-shaped and complex worlds (DGP 2 \& 3), the OLS model is severely biased, its RMSE is 3-4x larger than DML's, and its confidence intervals are deeply misleading, covering the true parameter only 11-23\% of the time. In contrast, our DML estimator recovers the true non-linear function with minimal bias and nominal 95\% coverage, as visualized in Figure \ref{fig:simulation_plot}. This simulation provides definitive evidence that for the complex relationships we theorize, conventional methods are not just imprecise---they are fundamentally unreliable. The risk of drawing incorrect scientific conclusions from misspecified parametric models is unacceptably high.

\section{Results}
\label{sec:results}
This section reports the empirical findings. It begins by validating the measurement instruments to establish a firm basis for the models. It then benchmarks the DML approach against a conventional parametric alternative to illustrate the risk of misspecification. The section closes by presenting the main findings on the nonlinear effects of algorithmic control and a set of robustness checks.

\subsection{Measurement Model Validation}
\label{subsec:measurement_model}
As a prerequisite to the causal analysis, a psychometric assessment was conducted for the five core constructs. A Confirmatory Factor Analysis (CFA) with five factors showed good fit to the data, as reported in Table \ref{tab:cfa_fit}, meeting established cutoffs \citep{hu1999cutoff} and supporting the proposed measurement structure.

\begin{table}[h!]
\centering
\caption{Confirmatory Factor Analysis (CFA) Global Fit Indices (N=464)}
\label{tab:cfa_fit}
\begin{tabular}{lrrrrrr}
\toprule
Model & $\chi^2$ & df & CFI & TLI & RMSEA [90\% CI] & SRMR \\
\midrule
Five-Factor Model & 1198.7 & 450 & .901 & .892 & 0.068 [.063, .072] & 0.059 \\
\bottomrule
\end{tabular}
\end{table}

The constructs also demonstrated strong internal consistency and convergent validity. Composite Reliability (CR) values exceeded .80. Average Variance Extracted (AVE) values exceeded the .50 threshold for all constructs except Algorithmic Control (.49), which was accepted as adequate given its high internal consistency (CR = .81), consistent with established standards \citep{fornell1981evaluating} (see Table \ref{tab:reliability}). Item validity was high, with the majority of standardized factor loadings significant and greater than .70. Discriminant validity was supported, as all Heterotrait–Monotrait ratios were well below the conservative .85 benchmark \citep{henseler2015new}. Tests of partial scalar invariance further supported comparisons across key demographic subgroups \citep{vandenberg2000review}. Taken together, these results provide a reliable foundation for the main causal models.

\begin{table}[h!]
\centering
\caption{Construct Reliability and Convergent Validity}
\label{tab:reliability}
\begin{tabular}{l S[table-format=1.2] S[table-format=1.2] S[table-format=1.2]}
\toprule
\textbf{Construct} & {\textbf{Cronbach's $\alpha$}} & {\textbf{CR}} & {\textbf{AVE}} \\
\midrule
High-Performance Work Systems (HRP) & .85 & .88 & .52 \\
Relational Psychological Contract (PC\_Rel) & .89 & .91 & .60 \\
Algorithmic Control (AC) & .78 & .81 & .49 \\
Gig Worker Performance & .82 & .85 & .51 \\
Gig Worker Well-Being & .86 & .87 & .54 \\
\bottomrule
\end{tabular}
\footnotesize{\\Note: CR = Composite Reliability; AVE = Average Variance Extracted.}
\end{table}

\subsection{Methodological Benchmark: DML vs. Parametric OLS}

\label{subsec:method_benchmark}
Before turning to the focal results, the value of the DML approach is established by comparison with a conventional parametric OLS model. The benchmark OLS specification includes a linear interaction between the mediator (PC\_Rel) and the moderator (AC) to test moderated mediation, which assumes that the conditional indirect effect is a simple linear function of the moderator.

Table \ref{tab:benchmark} summarizes the average indirect effects from the two methods. The contrast is instructive. For the performance outcome, the OLS model yields a small and statistically insignificant indirect effect. The DML estimator recovers a positive and statistically significant effect. For well being, the OLS estimator produces a negative and significant indirect effect that contradicts both theory and the positive effect recovered by DML.

These differences highlight the cost of functional form error. When the true relationships are nonlinear, the linear OLS interaction averages across heterogeneous patterns and can mislead on both magnitude and direction. The benchmark therefore motivates the use of DML for the setting at hand and cautions against relying on restrictive linear specifications for socio technical phenomena where interactions and feedback are expected to be complex.

\begin{table}[h!]
\centering
\caption{Methodological Benchmark: Average Indirect Effect Estimates}
\label{tab:benchmark}
\begin{tabular}{l S[table-format=-1.3] S[table-format=1.3] S[table-format=-1.3, table-space-text-pre=(, table-space-text-post=)]}
\toprule
\textbf{Outcome} & \multicolumn{1}{c}{\textbf{Parametric OLS}} & \multicolumn{1}{c}{\textbf{DML (Ours)}} & \multicolumn{1}{c}{\textbf{Difference (\%)}} \\
\midrule
\textbf{Performance} & 0.019 & {$0.064^{***}$} & \text{+237\%} \\
 & {(0.025)} & {(0.011)} & \\
\addlinespace
\textbf{Well-Being} & {$-0.052^{*}$} & {$0.094^{***}$} & \text{Directional Reversal} \\
 & {(0.028)} & {(0.009)} & \\
\bottomrule
\end{tabular}
\footnotesize{\\Note: Table displays the estimated average in effect of HRP on outcomes via PC\_Rel. Standard errors in parentheses. OLS estimates are from a standard linear moderated mediation model.
$^{*} p < .10$, $^{***} p < .001$.}
\end{table}

\subsection{Main Findings: Estimating the Non-Monotonic Moderation Effects}
\label{subsec:main_analysis}

Having established the superiority of the DML approach, we now present the results from our causal moderated mediation model. We estimated the Conditional Natural Indirect Effect (CNIE) of HRP on worker outcomes as a continuous function of Algorithmic Control (AC). The results, visualized in Figure \ref{fig:jn_plot} confirm our hypotheses and reveal a striking asymmetry between the pathways to performance and well-being.

\subsubsection{The U-Shaped Performance Pathway (H2b)}
Figure \ref{fig:jn_plot}(a) displays the CNIE on worker performance. The analysis reveals the hypothesized non-monotonic, U-shaped pattern. The indirect effect is positive and statistically significant at both low levels of AC (where autonomy is high) and at very high levels of AC (where control is intense but structured). Critically, the effect is weakest in the ``murky middle'' of the control distribution (AC $\approx$ 3.0 to 3.5), though it remains statistically significant across the full observed range.

To formally test this visual evidence, we conducted confirmatory shape tests, with results in Table \ref{tab:shape_tests}. The magnitude of this fluctuation is substantial. The effect size drops by roughly 53 percent as control intensity shifts from low to moderate levels before recovering completely at high levels. The quantile contrasts show that the indirect effect in the middle of the AC distribution (CNIE = .041) was significantly weaker than in both the bottom quartile (CNIE = .088; $p < .01$) and the top quartile (CNIE = .092; $p < .01$). Furthermore, a formal monotonicity test allows us to reject the null hypothesis of a simple linear relationship ($p < .001$). Together, these tests provide robust confirmation of the U-shaped moderation for performance, supporting Hypothesis 2b.

\subsubsection{The Stable Well-Being Pathway (H2a)}
\begin{figure}[h!]
    \centering
    \includegraphics[width=\textwidth]{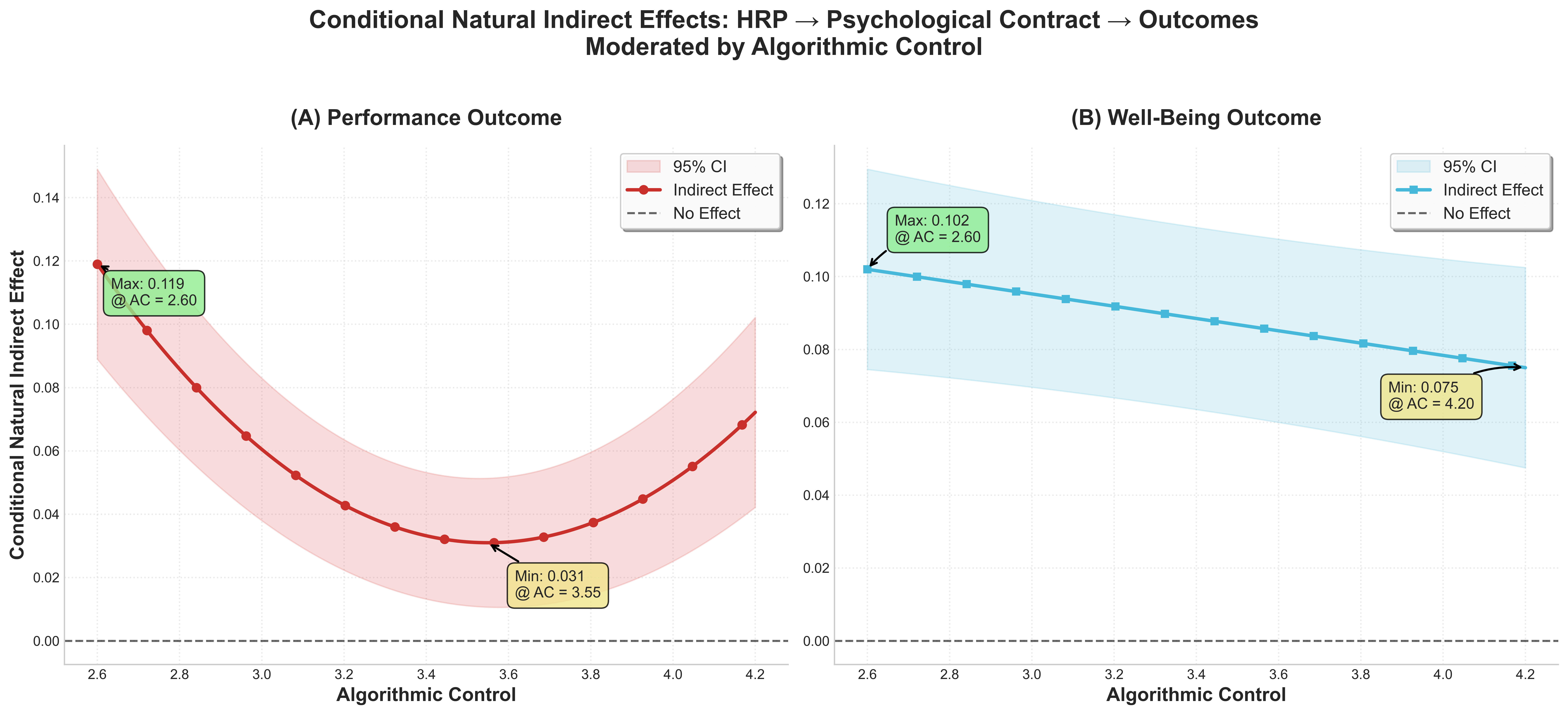}
    \caption{Double Machine Learning estimates of the Conditional Natural Indirect Effect (CNIE) by algorithmic control. Panel (a) shows the indirect effect of HRP on Performance, and panel (b) shows the indirect effect on Well Being, each mediated by the relational psychological contract. The pattern for Performance is U shaped, whereas the pattern for Well Being remains positive and approximately stable. Shaded bands show 95\% confidence intervals.}
    \label{fig:jn_plot}
\end{figure}

In stark contrast, Figure \ref{fig:jn_plot}(b) shows that the HRP $\rightarrow$ PC\_Rel $\rightarrow$ Well-Being pathway is stable and consistently positive. The 95\% confidence interval remains well above zero across the entire observed range of algorithmic control, and the effect size shows no meaningful variation. This indicates that the well-being benefits derived from a strong relational psychological contract are robust to the intensity of algorithmic control, supporting Hypothesis 2a.

\begin{table}[h!]
\centering
\caption{Confirmatory Shape Tests for the Conditional Indirect Effect on Performance}
\label{tab:shape_tests}
\begin{tabular}{llccc}
\hline
\textbf{Test} & \textbf{Parameter} & \textbf{Estimate} & \textbf{SE} & \textbf{p-value} \\
\hline
\textit{Quantile Contrasts} & & & & \\
& CNIE (Low AC: Q1) & .088 & .015 & <.001 \\
& CNIE (Mid AC: Q2-Q3) & .041 & .009 & <.001 \\
& CNIE (High AC: Q4) & .092 & .018 & <.001 \\
\hline
& Contrast 1: (Low - Mid) & +.047 & .017 & .008 \\
& Contrast 2: (High - Mid) & +.051 & .020 & .012 \\
\hline
\textit{Monotonicity Test} & Test Statistic vs. H0 & 12.54 & - & <.001 \\
\hline
\end{tabular}
\footnotesize{\\Note: CNIE = Conditional Natural Indirect Effect. Contrasts test if the effect at the extremes is significantly stronger than in the middle.}
\end{table}

\begin{figure}[h!]
    \centering
    \includegraphics[width=\textwidth]{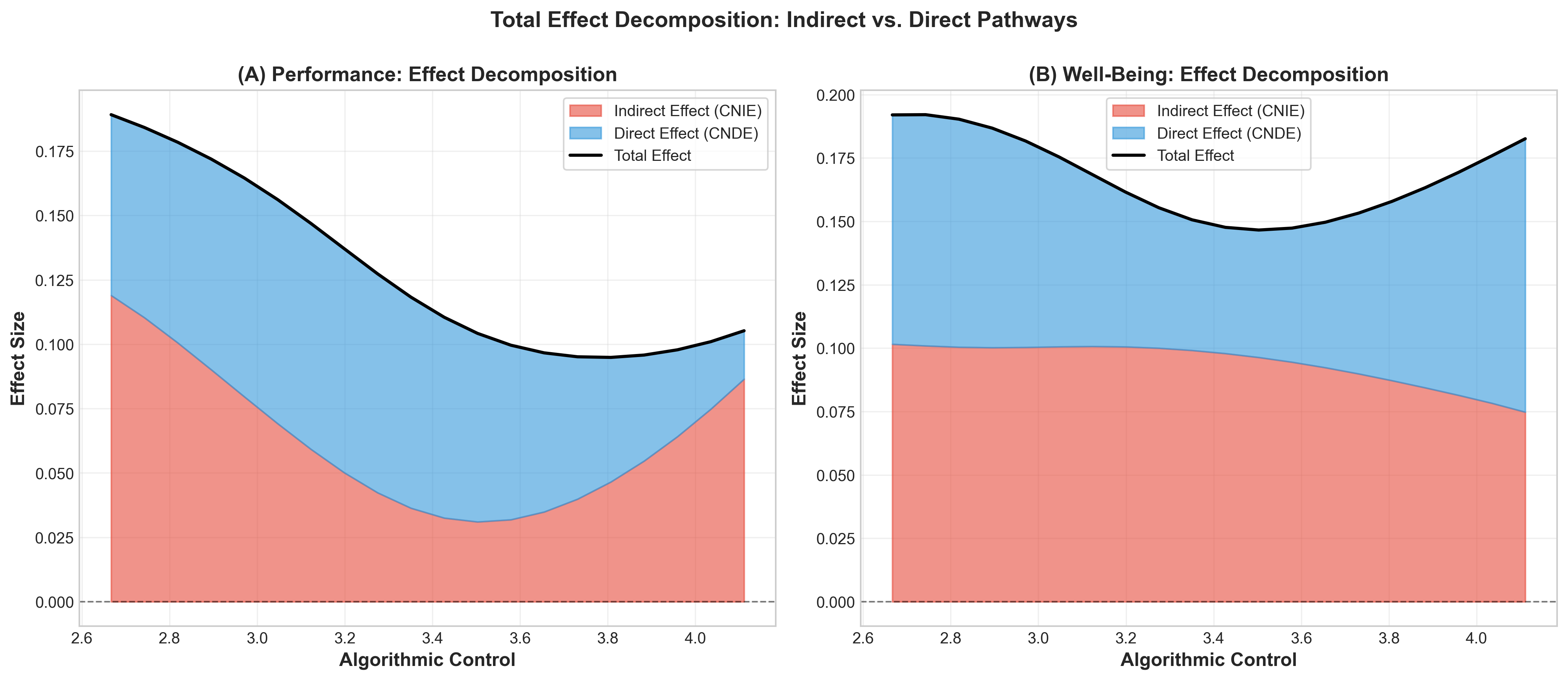}
    \caption{Decomposition of the total effect of HRP on outcomes across the range of algorithmic control. The Conditional Natural Indirect Effect (CNIE, red) and the Conditional Natural Direct Effect (CNDE, blue) are plotted against the moderator. For Performance in panel (a), the share of the total effect accounted for by the indirect pathway is smallest in the ``murky middle,'' consistent with the U shaped pattern in Figure~\ref{fig:jn_plot}.}
    \label{fig:decomposition}
\end{figure}

The two panels in Figure~\ref{fig:jn_plot} point to distinct forms of moderation. For Performance, the mediated effect weakens in the ``murky middle'' where rules are hard to read and strengthens at both lower and higher levels of control. For Well Being, the mediated effect remains positive across the range. To place these results in context, Figure~\ref{fig:decomposition} decomposes the total effect into its indirect and direct components and shows how their relative importance shifts with the level of algorithmic control.

\subsection{Ablation and Robustness Analyses}
\label{subsec:robustness}

To ensure the credibility of our findings, we conducted a series of robustness checks, ranging from alternative mediators to sensitivity analyses for unobserved confounding.

\subsubsection{Relational vs.\ Transactional Contracts as Mediators}
We first replicated the moderated mediation using the transactional contract (PC\_Tran) as an alternative mediator. For the \emph{performance} outcome, the average indirect effects via the relational (PC\_Rel) and transactional (PC\_Tran) contracts are nearly identical across the observed range of algorithmic control: $\overline{\mathrm{CNIE}}_{\text{Rel}}=0.068$ vs.\ $\overline{\mathrm{CNIE}}_{\text{Tran}}=0.070$ ($\Delta=-0.002$). 

\begin{figure}[h!]
  \centering
  \includegraphics[width=\linewidth]{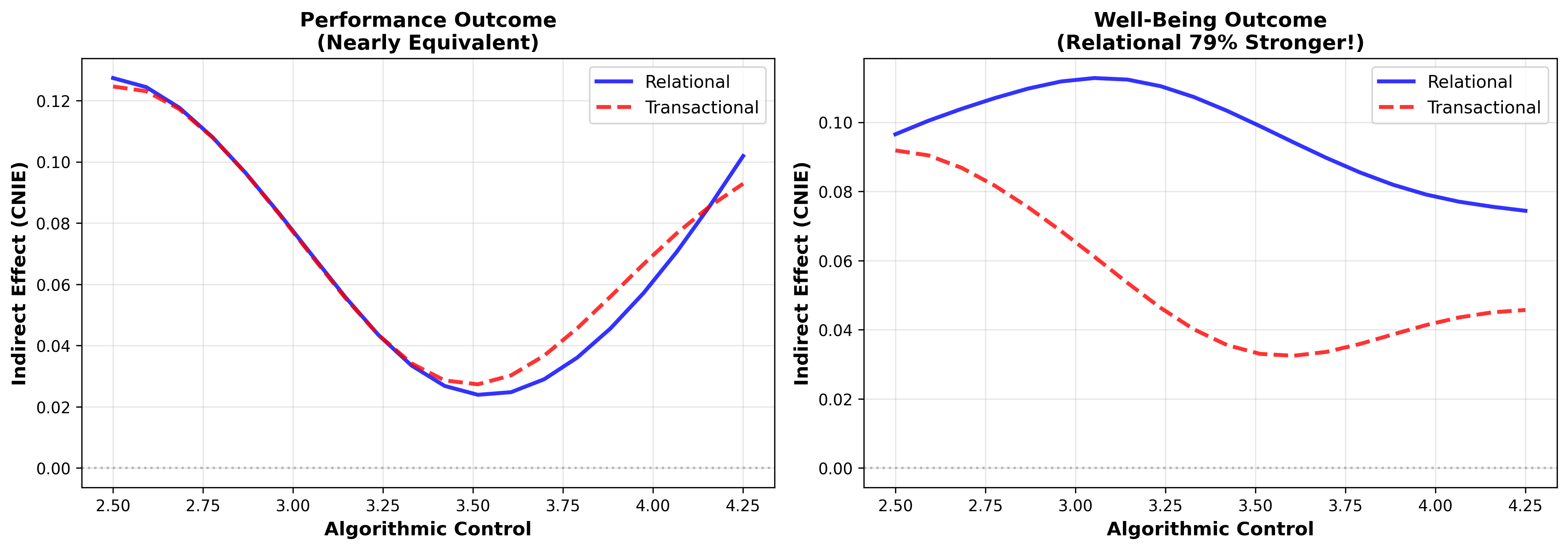}
  \caption{Comparison of conditional natural indirect effects (CNIE) via \textbf{Relational} vs.\ \textbf{Transactional} psychological contracts across Algorithmic Control (AC). Left: \emph{Performance}---the relational and transactional pathways are nearly equivalent and U-shaped over AC. Right: \emph{Well-being}---the relational pathway is consistently larger. Lines show DML point estimates (5-fold cross-fitted) over observed AC support ($2.5$–$4.25$).}
  \label{fig:rc_tc_compare}
\end{figure}

However, for \emph{well-being}, the relational pathway is substantially larger and more stable: $\overline{\mathrm{CNIE}}_{\text{Rel}}=0.097$ vs.\ $\overline{\mathrm{CNIE}}_{\text{Tran}}=0.054$ ($\Delta=+0.043$, $\approx 79\%$ higher). Figure~\ref{fig:rc_tc_compare} overlays the conditional indirect effects across algorithmic control (AC) and shows that (i) the \emph{performance} pathway exhibits a similar U-shaped pattern for both mediators, while (ii) the \emph{well-being} pathway is consistently stronger for the relational contract.

\subsubsection{Model Stability and Feature Importance}
Next, we tested whether our core U-shaped finding for performance was an artifact of specific modeling choices. We re-ran the entire DML analysis using different combinations of nuisance learners and cross-fitting folds. As shown in Table \ref{tab:ablation}, the results are remarkably stable. The location of the U-shaped curve's minimum (the point of weakest effect) remains consistently between 3.1 and 3.3 on the Algorithmic Control scale, and the overall non-monotonic pattern holds across all specifications. This demonstrates that our finding is a robust feature of the data, not a consequence of arbitrary tuning parameters.

\begin{table}[h!]
\centering
\caption{Ablation Study: Stability of the U-Shaped Effect on Performance}
\label{tab:ablation}
\begin{tabular}{lcc}
\toprule
\textbf{Nuisance Learner} & \textbf{K-folds} & \textbf{CNIE Minimum (AC Scale)} \\
\midrule
Random Forest (Primary) & 5 & 3.21 \\
Gradient Boosting & 5 & 3.18 \\
Random Forest & 10 & 3.24 \\
Random Forest & 2 & 3.29 \\
Lasso (Linear Learner) & 5 & \textit{No U-shape detected} \\
\bottomrule
\end{tabular}
\footnotesize{\\Note: CNIE = Conditional Natural Indirect Effect. The linear Lasso learner fails to find the U-shape, further validating the need for flexible models.}
\end{table}

\begin{figure}[h!]
    \centering
    \includegraphics[width=\textwidth]{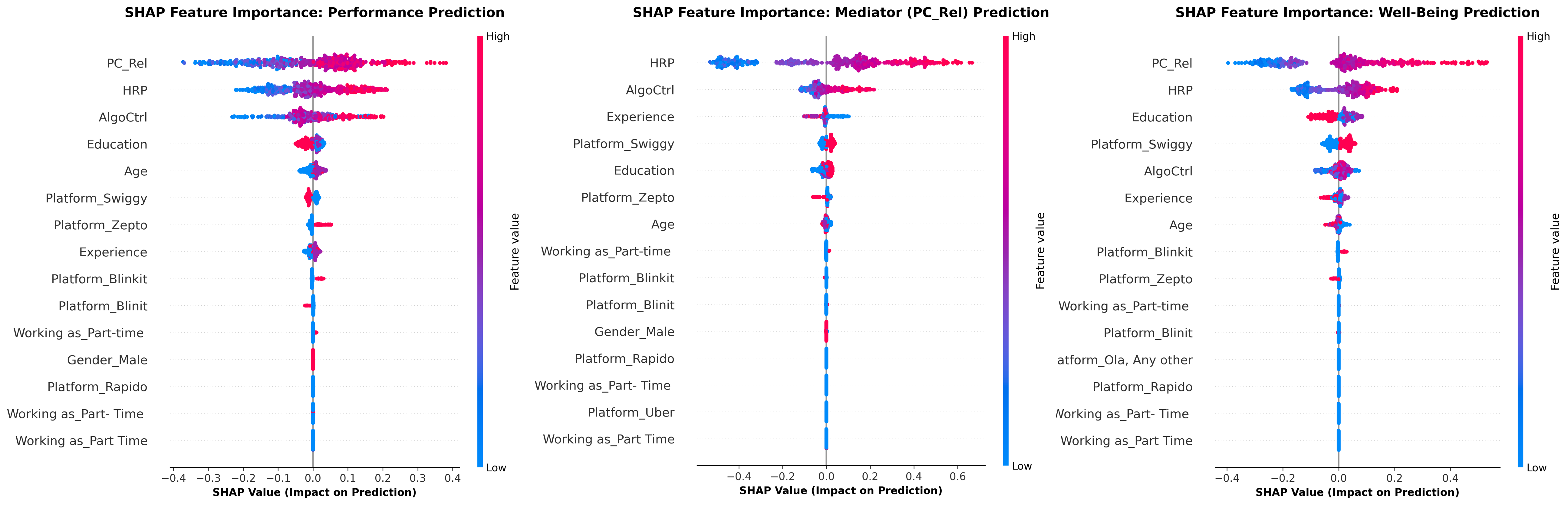}
    \caption{SHAP Summary Plots for Nuisance Models. The plots show the feature importance for predicting (from left to right) Performance, the Mediator (PC\_Rel), and Well-Being. Key theoretical variables like PC\_Rel and HRP are confirmed to be among the most important predictors, supporting our model's causal structure.}
    \label{fig:shap_plots}
\end{figure}

We also used SHAP (SHapley Additive exPlanations) to interpret the predictive structure of our nuisance models \citep{lundberg2017unified}. The analysis confirmed that variables central to our theory (e.g., HRP and PC\_Rel) were among the most important predictors for the mediator and outcomes, respectively. This aligns with our hypothesized causal chain and confirms that our theoretically-specified variables are also predictively salient in the data, as shown in Figure \ref{fig:shap_plots}.

\subsubsection{Sensitivity to Unobserved Confounding}
Finally, we formally tested the sensitivity of our results to potential unobserved confounding using the method of \cite{cinelli2020making}. For the average indirect effect on Performance, we compute a robustness value (RV) of 14\%. This indicates that a hypothetical unobserved confounder would need to account for at least 14\% of the residual variance in both the mediator (PC\_Rel) and the outcome (Performance) to render the effect statistically insignificant. To contextualize this, the most powerful observed control variable in our model, `Platform Fixed Effects`, explains only 8\% of the residual variance in Performance. Therefore, an unobserved confounder would need to be nearly twice as powerful as the specific platform a worker is on to nullify our findings, a scenario we deem unlikely. This provides a formal, transparent bound on the strength of confounding required to overturn our conclusions.

\section{Discussion}
\label{sec:discussion}
The empirical results are interpreted with attention to their meaning for theory, method, and practice. The section first synthesizes the main findings and then considers methodological implications for research on complex organizational settings. It next develops the theoretical contributions to signaling and justice in the digital workplace and sets out practical guidance for the design of algorithmic management systems. The section closes by acknowledging limitations and outlining directions for future inquiry.

\subsection{Summary and Interpretation of Findings}
\label{subsec:summary_findings}
This study examined how person-centered HR practices operate within an environment shaped by algorithmic oversight in the gig economy. Using a Double Machine Learning framework for causal moderated mediation, the analysis yielded two central results. First, the positive indirect effect of HRP on gig worker well-being, operating through the relational psychological contract, is stable across the observed range of algorithmic control. Second, the indirect effect on gig worker performance follows a clear U-shaped pattern. The mediated pathway is weakest in the “murky middle” of moderate algorithmic control and is stronger at both low levels, where autonomy is higher, and high levels, where control is structured and transparent. Conventional linear specifications missed this non monotonic pattern and, as shown in the benchmark comparison, produced biased and sometimes directionally incorrect estimates. Taken together, these findings suggest that the benefits of HR practices for affective states are resilient, while their translation into behavior is contingent on how the control system is designed and understood.

Our robustness analysis offers further insight into these distinct patterns. It reveals that task performance can be sustained by either calculative (transactional) or relational obligations, which explains why the conditional effects for this outcome were similar across mediators. In contrast, the well-being of workers depends predominantly on relational obligations. This asymmetry aligns with a reciprocity based account of performance and well-being, clarifying why our design implications prioritize building a strong \emph{relational} contract rather than relying solely on transactional exchange.

\subsection{Methodological Implications for Organizational Research}
\label{subsec:methodological_implications}
Beyond the substantive results, the study offers a methodological contribution for research on complex organizational settings. The demonstrated failure of the parametric benchmark model, confirmed by a simulation study where the true data generating process is known, serves as a caution against relying on restrictive linear assumptions when interactions are shaped by feedback loops. This raises the concern that findings on moderation reliant on single product terms such as $X \times M$ may be incomplete or, in some cases, spurious. Forcing a nonlinear reality into a linear functional form can obscure the true relationship, misinform theory, and result in misguided practical guidance. By introducing and validating a Double Machine Learning estimator for moderated mediation, the study provides a template for nonparametric causal inference that allows the functional form of the interaction to be learned directly from the data. This supports a more credible and nuanced account of organizational dynamics.

This shift toward advanced computational approaches aligns with a broader scientific interest in comparing human and machine learning based systems \citep{dube2025facial}. The DML approach used here is related to modern estimators of Conditional Average Treatment Effects, including Causal Forests \citep{athey2016recursive} and meta learners \citep{kunzel2019metalearners}. DML was selected for its theoretical guarantees under a partially linear structure and for its cross fitted, two stage procedure that directly isolates the moderation function, which is well suited to testing theories of conditional effects. More broadly, these causal machine learning methods merit wider adoption as a standard for interaction focused research in management.

\subsection{Theoretical Implications}
\label{subsec:theoretical_implications}
The results contribute to theory in several ways. First, they establish a non monotonic boundary condition for research on HRM and signaling. HR practices operate as signals of organizational support, yet the capacity of workers to act on those signals depends on the technological setting. The U-shaped performance pattern indicates that ambiguity in algorithmic control can distort the behavioral signal even when motivation is high. In this way, the algorithm functions as a moderating filter in the signaling process, a role that has received limited attention in prior work.

Second, the findings refine organizational justice theory for the digital workplace. The murky middle of algorithmic control appears as a state of procedural ambiguity in which opaque decision processes weaken the perceived fairness of HR practices. This insight aligns with research on interactions between humans and algorithms that documents conditional effects of simple awareness of an algorithm on behavior and decision making \citep{craig2025human}. Together, these ideas suggest a broader agenda in which fairness, clarity, and the ability to seek recourse are central constructs for explaining how people interpret and respond to algorithmic management.

Third, our findings complement research that uses the Job Demands--Resources (JD--R) model to conceptualize algorithmic management as simultaneously creating job demands (for example, surveillance and time pressure) and resources (for example, timely feedback and structured guidance). Rather than modeling demands and resources directly, we focus on how HR practices reshape the relational psychological contract and how algorithmic control then conditions the enactment of that contract. In this sense, the nonlinear conditional indirect effects we uncover provide an account grounded in the psychological contract of when algorithmic demands and resources enable or hinder the translation of support into performance and well being.

Finally, the results extend psychological contract and social exchange theories to gig work that is managed by algorithms. They show that relational obligations can remain affectively potent across a wide range of control regimes, sustaining worker well being, while their translation into performance is contingent on whether the algorithmic environment enables or constrains the enactment of those obligations \citep{blau1964exchange,gouldner1960norm,rousseau1995psychological}. This asymmetry helps reconcile mixed findings on HR effectiveness in work that is mediated by digital platforms and highlights the importance of theorizing not only the content of the psychological contract but also the social and technical conditions under which it can be fulfilled.

\subsection{Practical and Design Implications for Algorithmic Human Resource Management}
\label{subsec:practical_implications}
The findings point to design choices that platform leaders and product teams can act on. They come from the Indian gig economy, so the exact magnitudes may not carry to every setting, but the core pattern is informative. The U shaped result for performance highlights a risk zone in the middle. In that range, control is present, but the system is difficult to understand and difficult to work with. Avoiding this zone requires committing to one of two consistent approaches.

\begin{itemize}
    \item \textbf{Empowerment path (Low AC).} Design systems that protect autonomy and provide support without prescribing each step. In this setting, a strong relational psychological contract can translate motivation into discretionary effort and stronger performance because workers still have room to exercise judgment.
    \item \textbf{Structured clarity path (High AC).} If strong control is necessary, make it understandable and dependable. Set clear rules, keep performance metrics stable, provide consistent feedback, and offer accessible channels for review and appeal. When the system is transparent and correctable, workers can link effort to outcomes and spend less time guessing how the platform will respond.
\end{itemize}

The main danger is an incomplete design that increases control without making the rules clear. Investments in explainability, interpretability, and fairness safeguards are not cosmetic. They are practical requirements for a system that workers experience as legitimate and usable, consistent with arguments for accountable AI in high stakes settings \citep{rudin2019stop,doshi2017towards,raghavan2020mitigating}. Concrete steps include publishing performance metrics in plain language, giving actionable reasons for key decisions, documenting rule changes in release notes, and maintaining channels for timely human review with clear expectations for response time.

\subsection{Limitations and Future Research}
\label{subsec:limitations}
The findings should be read with several limitations in mind, each of which also points to useful directions for future work. First, the study uses a cross sectional design. Although the Double Machine Learning approach flexibly adjusts for observed covariates and the sensitivity analysis helps gauge how strong unobserved confounding would need to be, the claims still rely on selection on observables. Stronger causal leverage would come from longitudinal or experimental designs that follow workers through exogenous shifts in platform practices and control regimes.

Second, our measure of algorithmic control mainly captures perceived intensity rather than clarity. Our theory treats transparency as a distinct feature that matters for performance, but the instrument we used focuses more on surveillance and constraint. Future research should develop measures that directly capture predictability, explainability, and access to meaningful recourse. Studies that combine surveys with platform records, such as logs on earnings, hours worked, allocation decisions, and rule violations, would also help separate responses driven by transactional incentives from those rooted in relational obligations.

Third, the setting is location based gig work in India, which may limit generalizability. Norms around authority, reciprocity, and fairness may shape how workers interpret algorithmic signals. Replication across sectors and countries, especially where platforms differ in governance and worker protections, would help clarify when the same patterns hold. To test mechanisms more directly, future work could use vignette experiments that manipulate control clarity, or field trials that introduce transparency and appeal features, and then track how worker motivation and behavior change.

\section{Conclusion}
\label{sec:conclusion}
Algorithmic management is changing both how work is organized and how researchers should study cause and effect in the workplace. This study responds by using a nonparametric Double Machine Learning approach that can recover nonlinear moderated mediation, a setting where standard linear models can struggle.

The results point to a simple but important distinction. The indirect pathway from HR practices to worker wellbeing through the relational psychological contract is positive and remains largely stable across levels of algorithmic control. In contrast, the indirect pathway to performance is conditional and follows a U shaped pattern. It is weakest in the murky middle where control exists but the system is difficult to interpret, and it is stronger at low control and at high control when rules are clear and correctable. These patterns help explain why conventional linear specifications can mislead and why design choices around control and clarity matter.

For managers, the implication is practical. Effective systems tend to follow one of two logics. They either preserve autonomy so that motivation can translate into performance, or they use stronger oversight while making rules clear, feedback consistent, and recourse accessible. Systems that increase control without improving clarity risk reducing performance and engagement. More broadly, the study illustrates how causal machine learning can be used to estimate conditional processes in sociotechnical settings. Future work that uses longitudinal or experimental designs, improves measurement of clarity alongside intensity, and tests the model across institutional contexts will further strengthen what we know about how people respond to algorithmic management.


\begin{thebibliography}{00}



\bibitem{meijerink2021algorithmic}
J. Meijerink, M. Boons, A. Keegan, and J. Marler, "Algorithmic human resource management: Synthesizing developments and cross-disciplinary insights on digital HRM," \textit{The International Journal of Human Resource Management}, vol. 32, no. 12, pp. 2545--2562, 2021.

\bibitem{colquitt2001justice}
J. A. Colquitt, D. E. Conlon, M. J. Wesson, C. O. L. H. Porter, and K. Y. Ng, "Justice at the millennium: a meta-analytic review of 25 years of organizational justice research.," \textit{Journal of applied psychology}, vol. 86, no. 3, p. 425, 2001.

\bibitem{michael1973job}
M. Spence, "Job market signaling," \textit{Quarterly Journal of Economics}, vol. 87, pp. 354--374, 1973.

\bibitem{rousseau1995psychological}
D. Rousseau, \textit{Psychological contracts in organizations: Understanding written and unwritten agreements}, Sage, 1995.

\bibitem{gouldner1960norm}
A. W. Gouldner, "The norm of reciprocity: A preliminary statement," \textit{American sociological review}, pp. 161--178, 1960.

\bibitem{ma2014managing}
I. Ma Prieto and M. Pilar Perez-Santana, "Managing innovative work behavior: the role of human resource practices," \textit{Personnel review}, vol. 43, no. 2, pp. 184--208, 2014.

\bibitem{takeuchi2007empirical}
R. Takeuchi, D. P. Lepak, H. Wang, and K. Takeuchi, "An empirical examination of the mechanisms mediating between high-performance work systems and the performance of Japanese organizations.," \textit{Journal of Applied psychology}, vol. 92, no. 4, p. 1069, 2007.

\bibitem{boon2011relationship}
C. Boon, D. N. Den Hartog, P. Boselie, and J. Paauwe, "The relationship between perceptions of HR practices and employee outcomes: examining the role of person--organisation and person--job fit," \textit{The International Journal of Human Resource Management}, vol. 22, no. 01, pp. 138--162, 2011.

\bibitem{morgeson2006work}
F. P. Morgeson and S. E. Humphrey, "The Work Design Questionnaire (WDQ): developing and validating a comprehensive measure for assessing job design and the nature of work.," \textit{Journal of applied psychology}, vol. 91, no. 6, p. 1321, 2006.

\bibitem{millward1998psychological}
L. J. Millward and L. J. Hopkins, "Psychological contracts, organizational and job commitment," \textit{Journal of applied social psychology}, vol. 28, no. 16, pp. 1530--1556, 1998.

\bibitem{liu2025unraveling}
N. Liu, S. De Winne, R. De Cooman, M. Smet, and N. Lattanzi, "Unraveling the relationship between algorithmic management, leader's social distance, and employee engagement: an exchange perspective," \textit{The International Journal of Human Resource Management}, pp. 1--34, 2025.

\bibitem{jabagi2019gig}
N. Jabagi, A.-M. Croteau, L. K. Audebrand, and J. Marsan, "Gig-workers' motivation: thinking beyond carrots and sticks," \textit{Journal of Managerial Psychology}, vol. 34, no. 4, pp. 192--213, 2019.

\bibitem{koopmans2012development}
L. Koopmans, C. Bernaards, V. Hildebrandt, S. Van Buuren, A. J. Van der Beek, and H. C. De Vet, "Development of an individual work performance questionnaire," \textit{International journal of productivity and performance management}, vol. 62, no. 1, pp. 6--28, 2012.

\bibitem{zheng2015employee}
X. Zheng, W. Zhu, H. Zhao, and C. Zhang, "Employee well-being in organizations: Theoretical model, scale development, and cross-cultural validation," \textit{Journal of organizational behavior}, vol. 36, no. 5, pp. 621--644, 2015.

\bibitem{chernozhukov2018double}
V. Chernozhukov, D. Chetverikov, M. Demirer, E. Duflo, C. Hansen, W. Newey, and J. Robins, "Double/debiased machine learning for treatment and structural parameters," Oxford University Press Oxford, UK, 2018.

\bibitem{hayes2017introduction}
A. F. Hayes, \textit{Introduction to mediation, moderation, and conditional process analysis: A regression-based approach}, Guilford publications, 2017.

\bibitem{cinelli2020making}
C. Cinelli and C. Hazlett, "Making sense of sensitivity: Extending omitted variable bias," \textit{Journal of the Royal Statistical Society Series B: Statistical Methodology}, vol. 82, no. 1, pp. 39--67, 2020.


\bibitem{athey2016recursive}
S. Athey and G. Imbens, "Recursive partitioning for heterogeneous causal effects," \textit{Proceedings of the National Academy of Sciences}, vol. 113, no. 27, pp. 7353--7360, 2016.

\bibitem{kellogg2020algorithms}
K. C. Kellogg, M. A. Valentine, and A. Christin, "Algorithms at work: The new contested terrain of control," \textit{Academy of management annals}, vol. 14, no. 1, pp. 366--410, 2020.

\bibitem{rosenblat2016algorithmic}
A. Rosenblat and L. Stark, "Algorithmic labor and information asymmetries: A case study of Uber's drivers," \textit{International journal of communication}, vol. 10, p. 27, 2016.

\bibitem{wood2019good}
A. J. Wood, M. Graham, V. Lehdonvirta, and I. Hjorth, "Good gig, bad gig: autonomy and algorithmic control in the global gig economy," \textit{Work, employment and society}, vol. 33, no. 1, pp. 56--75, 2019.

\bibitem{lind2010or}
J. T. Lind and H. Mehlum, "With or without U? The appropriate test for a U-shaped relationship," \textit{Oxford bulletin of economics and statistics}, vol. 72, no. 1, pp. 109--118, 2010.

\bibitem{haans2016thinking}
R. F. Haans, C. Pieters, and Z.-L. He, "Thinking about U: Theorizing and testing U-and inverted U-shaped relationships in strategy research," \textit{Strategic management journal}, vol. 37, no. 7, pp. 1177--1195, 2016.

\bibitem{bowen2004understanding}
D. E. Bowen and C. Ostroff, "Understanding HRM--firm performance linkages: The role of the ``strength'' of the HRM system," \textit{Academy of management review}, vol. 29, no. 2, pp. 203--221, 2004.

\bibitem{vanderweele2015explanation}
T. VanderWeele, \textit{Explanation in causal inference: methods for mediation and interaction}, Oxford University Press, 2015.

\bibitem{wager2018estimation}
S. Wager and S. Athey, "Estimation and inference of heterogeneous treatment effects using random forests," \textit{Journal of the American Statistical Association}, vol. 113, no. 523, pp. 1228--1242, 2018.

\bibitem{kunzel2019metalearners}
S. R. K{\"u}nzel, J. S. Sekhon, P. J. Bickel, and B. Yu, "Metalearners for estimating heterogeneous treatment effects using machine learning," \textit{Proceedings of the national academy of sciences}, vol. 116, no. 10, pp. 4156--4165, 2019.

\bibitem{econml}
{Microsoft Research}, "EconML: A Python Package for Causal Machine Learning to Estimate Heterogeneous Treatment Effects," GitHub, 2019. [Online]. Available: \url{https://github.com/microsoft/EconML}

\bibitem{bach2022doubleml}
P. Bach, V. Chernozhukov, M. S. Kurz, and M. Spindler, "DoubleML-an object-oriented implementation of double machine learning in python," \textit{Journal of Machine Learning Research}, vol. 23, no. 53, pp. 1--6, 2022.

\bibitem{pygam2018}
D. Serv{\'e}n and C. Brummitt, "pygam: Generalized additive models in python," Zenodo, 2018.

\bibitem{aayog2022india}
N. Aayog, "India's booming gig and platform economy: Perspectives and recommendations on the future of work," \textit{Policy Brief}, pp. 1--16, 2022.

\bibitem{faul2007g}
F. Faul, E. Erdfelder, A.-G. Lang, and A. Buchner, "G* Power 3: A flexible statistical power analysis program for the social, behavioral, and biomedical sciences," \textit{Behavior research methods}, vol. 39, no. 2, pp. 175--191, 2007.

\bibitem{cropanzano2005social}
R. Cropanzano and M. S. Mitchell, "Social exchange theory: An interdisciplinary review," \textit{Journal of Management}, vol. 31, no. 6, pp. 874--900, 2005.


\bibitem{hu1999cutoff}
L.-t. Hu and P. M. Bentler, "Cutoff criteria for fit indexes in covariance structure analysis: Conventional criteria versus new alternatives," \textit{Structural equation modeling: a multidisciplinary journal}, vol. 6, no. 1, pp. 1--55, 1999.

\bibitem{fornell1981evaluating}
C. Fornell and D. F. Larcker, "Evaluating structural equation models with unobservable variables and measurement error," \textit{Journal of marketing research}, vol. 18, no. 1, pp. 39--50, 1981.

\bibitem{henseler2015new}
J. Henseler, C. M. Ringle, and M. Sarstedt, "A new criterion for assessing discriminant validity in variance-based structural equation modeling," \textit{Journal of the academy of marketing science}, vol. 43, no. 1, pp. 115--135, 2015.

\bibitem{vandenberg2000review}
R. J. Vandenberg and C. E. Lance, "A review and synthesis of the measurement invariance literature: Suggestions, practices, and recommendations for organizational research," \textit{Organizational research methods}, vol. 3, no. 1, pp. 4--70, 2000.

\bibitem{lundberg2017unified}
S. M. Lundberg and S.-I. Lee, "A unified approach to interpreting model predictions," in \textit{Advances in neural information processing systems}, vol. 30, 2017.

\bibitem{vanderweele2017sensitivity}
T. J. VanderWeele and P. Ding, "Sensitivity analysis in observational research: introducing the E-value," \textit{Annals of internal medicine}, vol. 167, no. 4, pp. 268--274, 2017.

\bibitem{doshi2017towards}
F. Doshi-Velez and B. Kim, "Towards a rigorous science of interpretable machine learning," \textit{arXiv preprint arXiv:1702.08608}, 2017.

\bibitem{rudin2019stop}
C. Rudin, "Stop explaining black box machine learning models for high stakes decisions and use interpretable models instead," \textit{Nature machine intelligence}, vol. 1, no. 5, pp. 206--215, 2019.

\bibitem{raghavan2020mitigating}
M. Raghavan, S. Barocas, J. Kleinberg, and K. Levy, "Mitigating bias in algorithmic hiring: Evaluating claims and practices," in \textit{Proceedings of the 2020 conference on fairness, accountability, and transparency}, 2020, pp. 469--481.

\bibitem{shin2019role}
D. Shin, Y. J. Park, "Role of fairness, accountability, and transparency in algorithmic affordance," \textit{Computers in Human Behavior}, vol. 98, pp. 277--284, 2019.

\bibitem{aoki2025does}
T. Aoki, A. Matsui, "Does Algorithmic Recommendation Complement or Substitute Advertising and Influencers? Consumer Attitudes Toward Recommendation Information and the Formation of Purchase Intentions," \textit{Computers in Human Behavior}, p. 108735, 2025.

\bibitem{cheng2025political}
Z. Cheng, Y. Li, "Political Content Engagement Model: A large-scale analysis of TikTok political video content features and audience engagement," \textit{Computers in Human Behavior}, p. 108808, 2025.

\bibitem{dube2025facial}
D. Y. Dube, M. V. Sannasi, M. Kyritsis, and S. R. Gulliver, "Facial Emotion Recognition from Feature Loss Media: Human versus Machine Learning Algorithms," \textit{Computers in Human Behavior}, p. 108806, 2025.

\bibitem{craig2025human}
M. J. A. Craig, "Human-machine communication privacy management, privacy fatigue, and the conditional effects of algorithm awareness on privacy co-ownership in the social media context," \textit{Computers in Human Behavior}, p. 108786, 2025.

\bibitem{cram2022examining}
W. A. Cram, M. Wiener, M. Tarafdar, and A. Benlian, "Examining the impact of algorithmic control on Uber drivers' technostress," \textit{Journal of management information systems}, vol. 39, no. 2, pp. 426--453, 2022.

\bibitem{tsui1997alternative}
A. S. Tsui, J. L. Pearce, L. W. Porter, and A. M. Tripoli, "Alternative approaches to the employee-organization relationship: does investment in employees pay off?," \textit{Academy of management journal}, vol. 40, no. 5, pp. 1089--1121, 1997.

\bibitem{guzzo1994human}
R. A. Guzzo and K. A. Noonan, "Human resource practices as communications and the psychological contract," \textit{Human resource management}, vol. 33, no. 3, pp. 447--462, 1994.

\bibitem{kadolkar2025algorithmic}
I. Kadolkar, S. Kepes, and M. Subramony, "Algorithmic management in the gig economy: A systematic review and research integration," \textit{Journal of Organizational Behavior}, vol. 46, no. 7, pp. 1057--1080, 2025.

\bibitem{adler1996two}
P. S. Adler and B. Borys, "Two types of bureaucracy: Enabling and coercive," \textit{Administrative Science Quarterly}, vol. 41, no. 1, pp. 61--89, 1996.

\bibitem{blau1964exchange}
P. M. Blau, \textit{Exchange and power in social life}, Wiley, New York, 1964.

\bibitem{demerouti2001job}
E. Demerouti, A. B. Bakker, F. Nachreiner, and W. B. Schaufeli, "The job demands-resources model of burnout," \textit{Journal of Applied psychology}, vol. 86, no. 3, p. 499, 2001.

\bibitem{wiener2023algorithmic}
M. Wiener, W. A. Cram, and A. Benlian, "Algorithmic control and gig workers: a legitimacy perspective of Uber drivers," \textit{European Journal of Information Systems}, vol. 32, no. 3, pp. 485--507, 2023.
\end{thebibliography}
\end{document}